\documentclass{article}
\usepackage{maniunit_preprint,times}
\usepackage{amsmath,amssymb}
\usepackage{booktabs,multirow,graphicx,array,tabularx}
\usepackage{xcolor}
\usepackage{url}

\usepackage{float,wrapfig,needspace}
\usepackage{placeins}

\graphicspath{{figures/}}
\newcommand{\papertitle}{ManiUnit: A Manipulation Skill Dataset\\and Benchmark for Long-Horizon Tasks}

\newenvironment{widefigure}{\begin{figure}}{\end{figure}}
\newenvironment{widetable}{\begin{table}}{\end{table}}
\definecolor{paperblue}{HTML}{2B536C}
\renewcommand{\arraystretch}{1.07}
\usepackage{hyperref}
\hypersetup{colorlinks=true,allcolors=paperblue,pdftitle={ManiUnit: A Manipulation Skill Dataset and Benchmark for Long-Horizon Tasks},pdfauthor={Guoting Wei, Dawei Yan, Xia Yuan, Gengming Zhang, Yelin He, Guodong Du, Jiaquan Ye, Heng Zhang, Xinming Wei, Xianbiao Qi, Chunxia Zhao, Haokui Zhang, Rong Xiao},pdfsubject={Robot manipulation skill dataset and benchmark}}
\title{\papertitle}
\author{%
  \begin{tabular}{@{}c@{}}
    \textbf{Guoting Wei}\textsuperscript{1,3}\thanks{Equal contribution.}\hspace{6pt}
    \textbf{Dawei Yan}\textsuperscript{2*}\hspace{6pt}
    \textbf{Xia Yuan}\textsuperscript{1*}\hspace{6pt}
    \textbf{Gengming Zhang}\textsuperscript{3}\hspace{6pt}
    \textbf{Yelin He}\textsuperscript{3}\hspace{6pt}
    \textbf{Guodong Du}\textsuperscript{3}\hspace{6pt}
    \textbf{Jiaquan Ye}\textsuperscript{3}\\[4pt]
    \textbf{Heng Zhang}\textsuperscript{1}\hspace{8pt}
    \textbf{Xinming Wei}\textsuperscript{3}\hspace{8pt}
    \textbf{Xianbiao Qi}\textsuperscript{3}\hspace{8pt}
    \textbf{Chunxia Zhao}\textsuperscript{1}\hspace{8pt}
    \textbf{Haokui Zhang}\textsuperscript{2,3}\thanks{Corresponding author.}\hspace{8pt}
    \textbf{Rong Xiao}\textsuperscript{3}\\[8pt]
    \textsuperscript{1}\,Nanjing University of Science and Technology\\[2pt]
    \textsuperscript{2}\,Northwestern Polytechnical University\\[2pt]
    \textsuperscript{3}\,Intellifusion
  \end{tabular}%
}

\newcommand{\FigureOverview}{%
\begin{widefigure}[H]
  \centering
  \includegraphics[width=\linewidth]{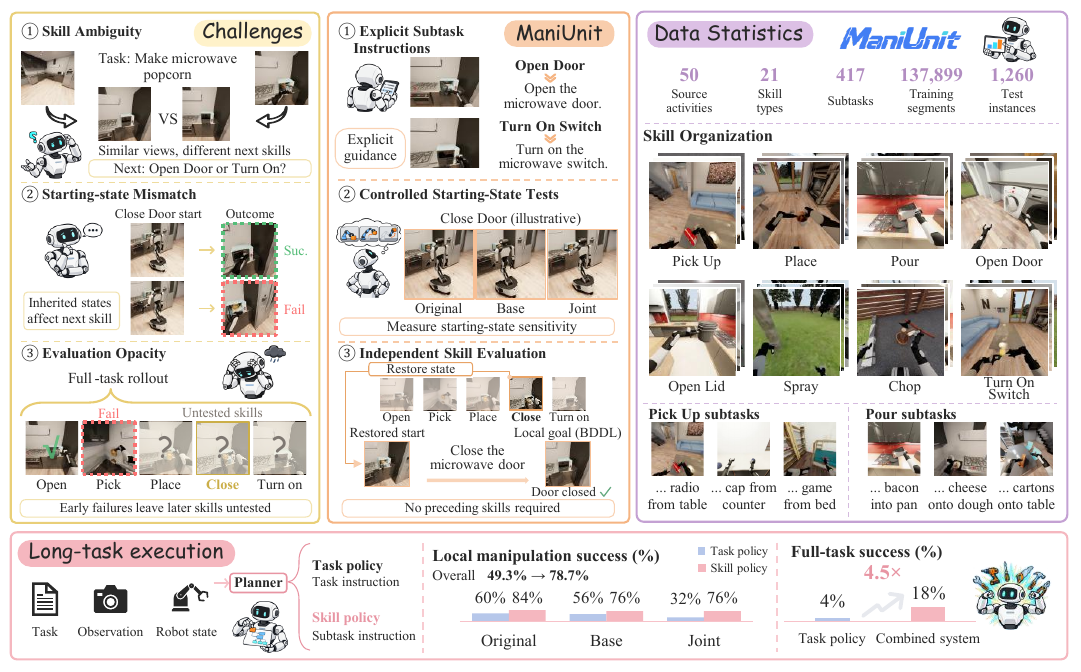}
  \caption{\textbf{ManiUnit overview.}
  \textbf{Left}: Long-horizon execution poses three challenges for skill learning
  and evaluation.
  \textbf{Center}: Correspondingly, ManiUnit provides explicit subtask instructions
  to guide manipulation. Controlled tests measure starting-state sensitivity,
  while restored states and local BDDL goals enable independent skill evaluation.
  \textbf{Right}: ManiUnit organizes manipulation segments by skill type and subtask.
  \mbox{\textbf{Bottom}:} In a focused study on two long-horizon activities,
  skill-level training improves local manipulation success, and combining the
  task and skill policies through a planner improves full-task success.}
  \label{fig:overview}
\end{widefigure}
}

\newcommand{\FigurePipeline}{%
\begin{widefigure}[t]
  \centering
  \includegraphics[width=\linewidth]{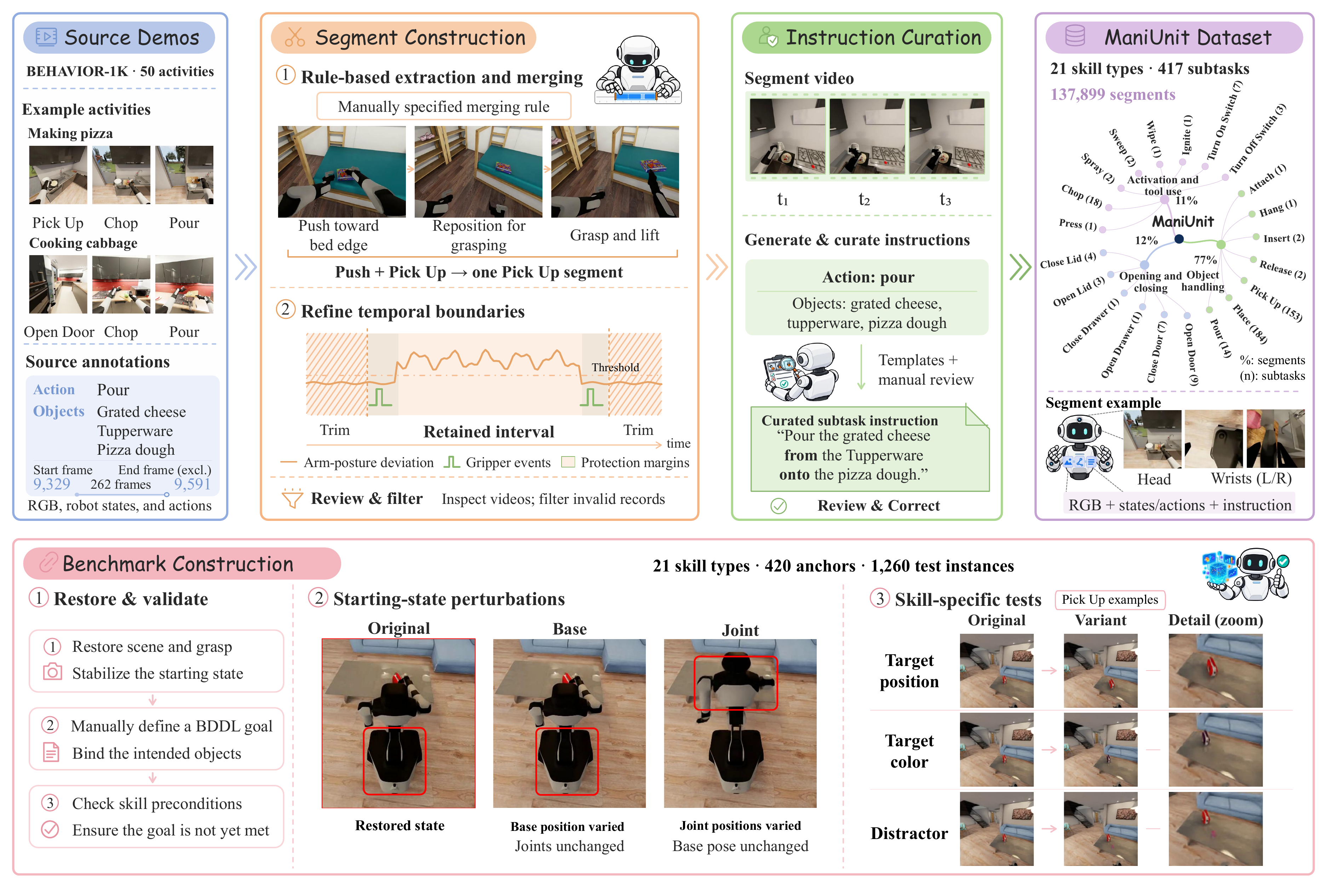}
  \caption{\textbf{ManiUnit dataset and benchmark construction.}
  Top: Source demonstrations are converted into manipulation segments
  with curated subtask instructions.
  Bottom: Each evaluation anchor marks a selected operation start, whose
  restored state, subtask instruction, and local BDDL goal enable
  independent skill evaluation across original and perturbed starting states.}
  \label{fig:pipeline}
\end{widefigure}
}

\newcommand{\FigurePlannerTrajectory}{%
\begin{widefigure}[!htbp]
  \centering
  \includegraphics[width=\linewidth]{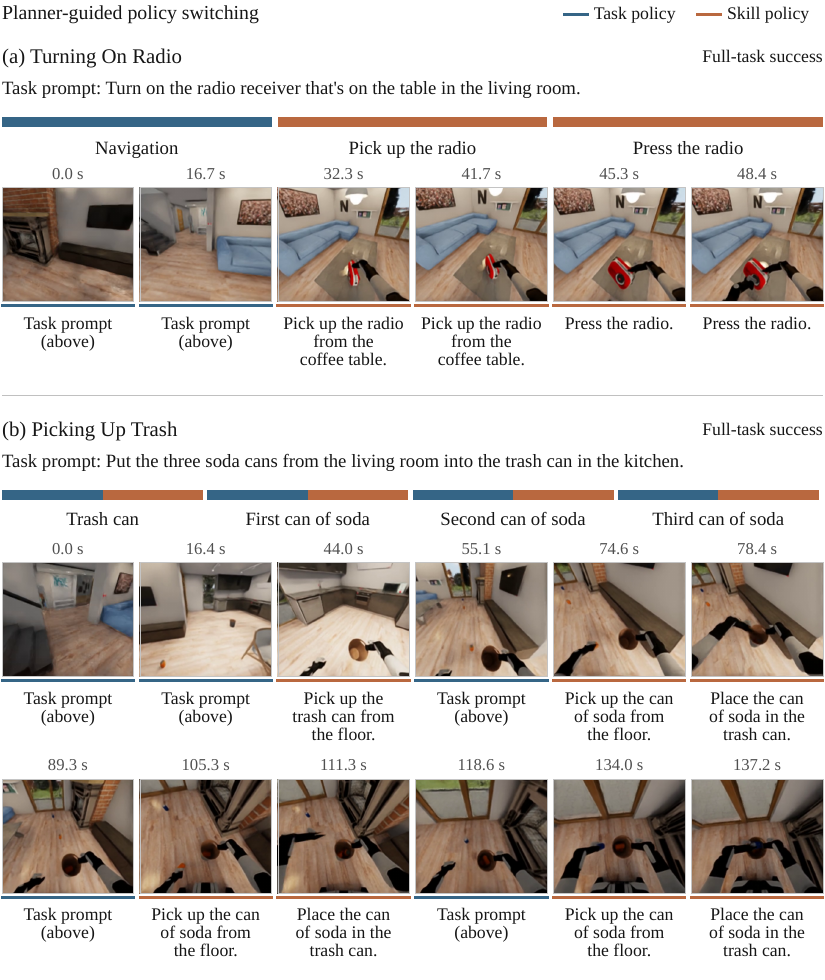}
  \caption{Successful rollouts of (a) Turning On Radio and (b) Picking Up Trash.
  A planner coordinates a task policy for navigation and task-level control
  with a ManiUnit-trained skill policy for manipulation.
  Colors indicate the active policy.}
  \label{fig:planner_trajectory}
\end{widefigure}
}

\newcommand{\FigureCatalogObjectsOne}{%
\begin{widefigure}[htbp]
\centering
\includegraphics[width=\linewidth,height=0.84\textheight,keepaspectratio]{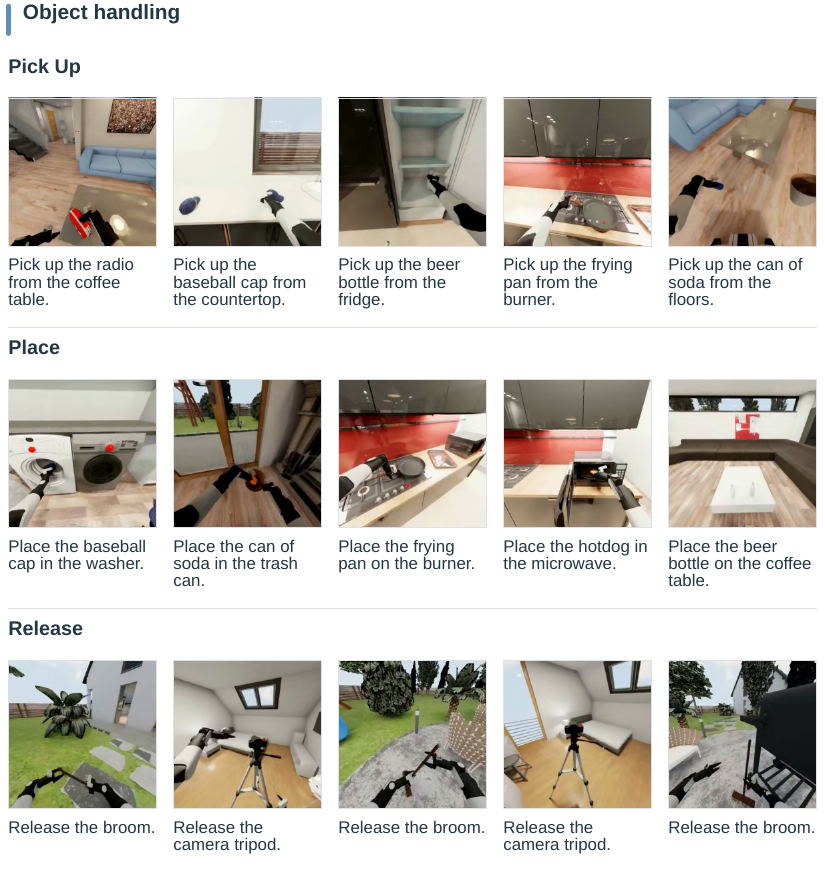}
\caption{Object handling: recorded examples with their subtask instructions.}
\label{fig:catalog_objects_one}
\end{widefigure}
}

\newcommand{\FigureCatalogObjectsTwo}{%
\begin{widefigure}[p]
\centering
\includegraphics[width=5.5in]{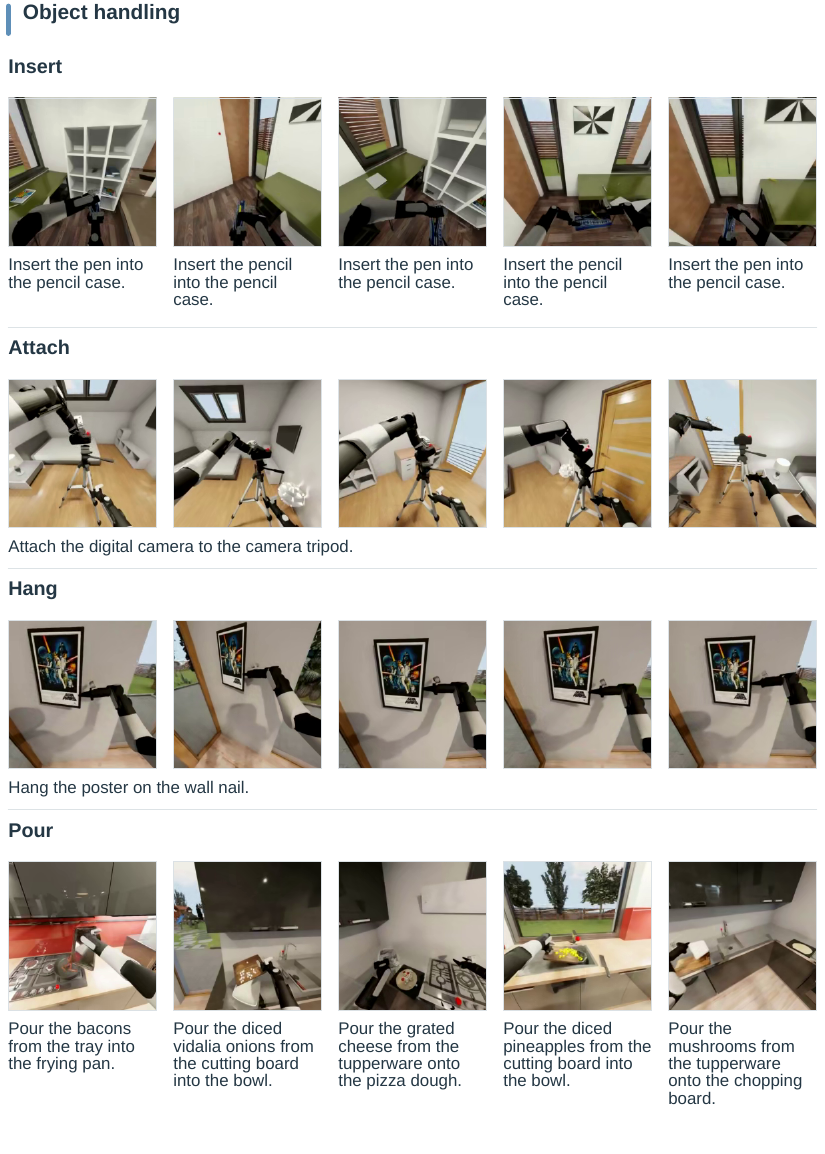}
\caption{Object handling: recorded examples with their subtask instructions.}
\label{fig:catalog_objects_two}
\end{widefigure}
}

\newcommand{\FigureCatalogArticulatedOne}{%
\begin{widefigure}[p]
\centering
\includegraphics[width=5.5in]{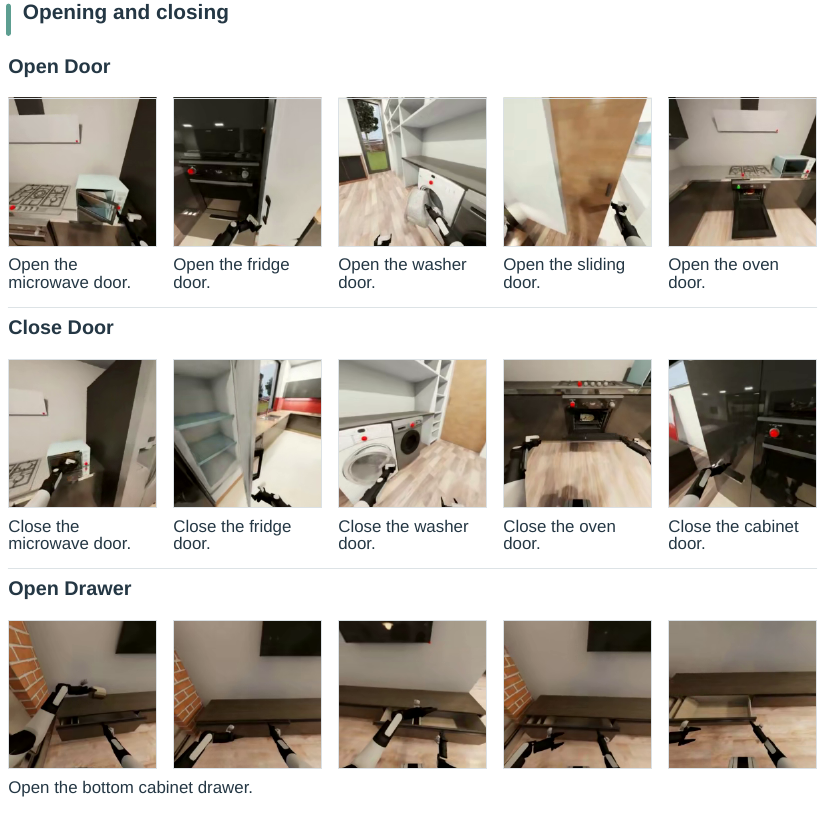}
\caption{Opening and closing: recorded examples with their subtask instructions.}
\label{fig:catalog_articulated_one}
\end{widefigure}
}

\newcommand{\FigureCatalogArticulatedTwo}{%
\begin{widefigure}[p]
\centering
\includegraphics[width=5.5in]{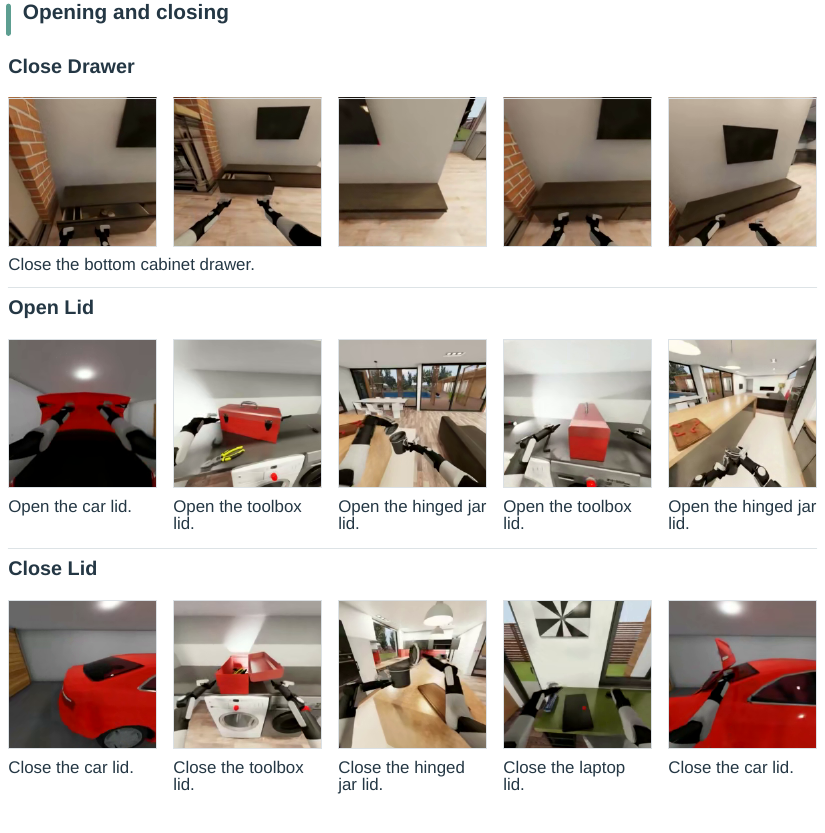}
\caption{Opening and closing: recorded examples with their subtask instructions.}
\label{fig:catalog_articulated_two}
\end{widefigure}
}

\newcommand{\FigureCatalogToolsOne}{%
\begin{widefigure}[p]
\centering
\includegraphics[width=5.5in]{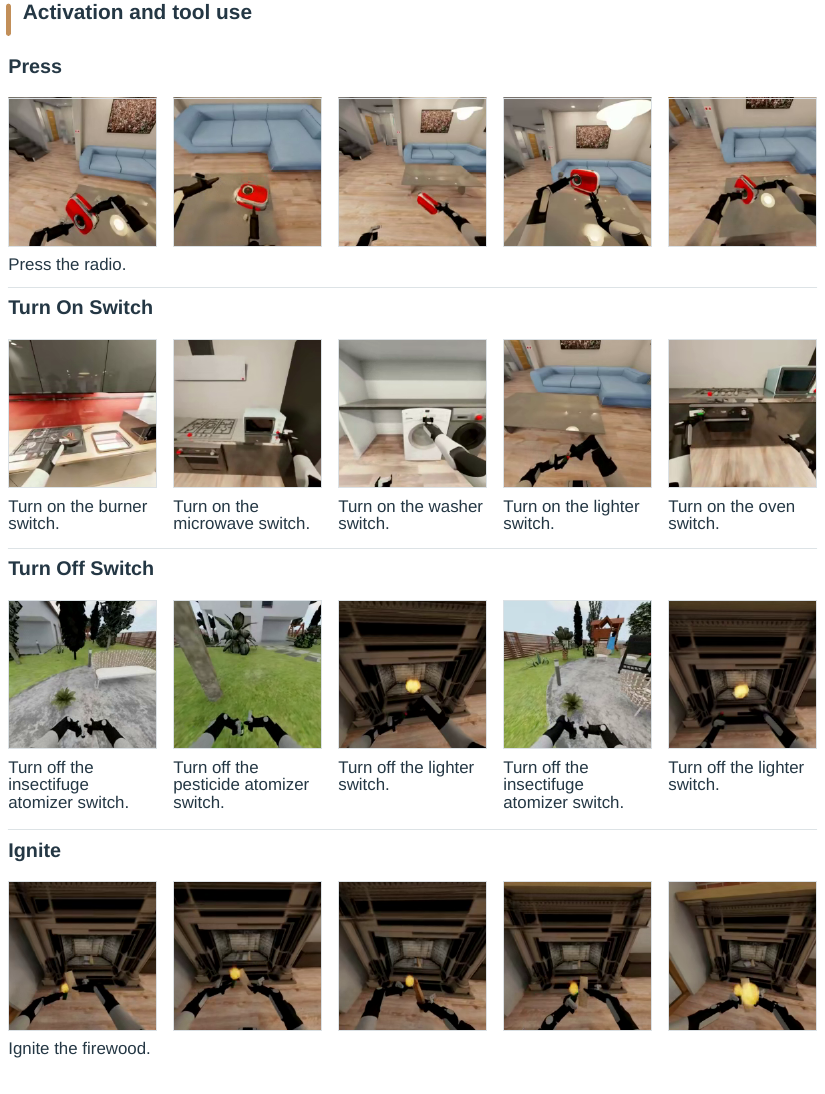}
\caption{Activation and tool use: recorded examples with their subtask instructions.}
\label{fig:catalog_tools_one}
\end{widefigure}
}

\newcommand{\FigureCatalogToolsTwo}{%
\begin{widefigure}[p]
\centering
\includegraphics[width=5.5in]{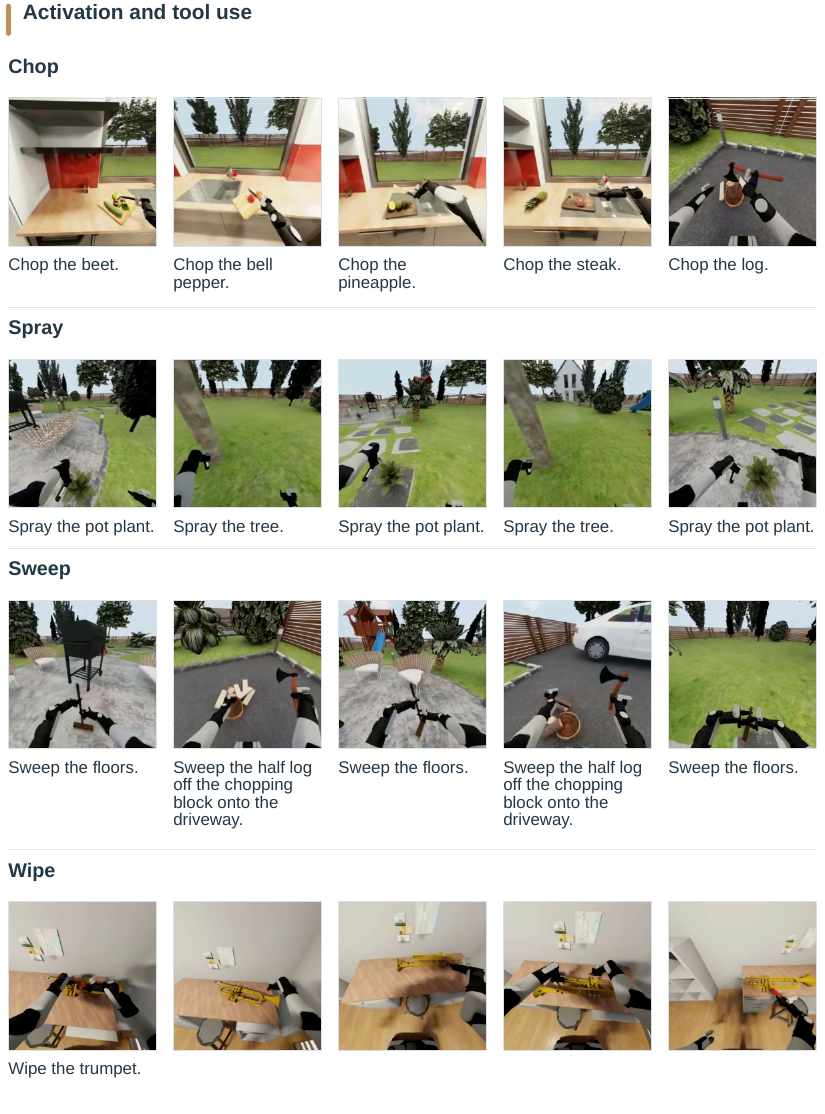}
\caption{Activation and tool use: recorded examples with their subtask instructions.}
\label{fig:catalog_tools_two}
\end{widefigure}
}

\newcommand{\ManiUnitFullTable}{%
\begin{tabular*}{\linewidth}{@{}l@{\extracolsep{\fill}}rrrr@{}}
\toprule
Model & Original & Base & Joint & Overall \\
\midrule
StarVLA-PI & 60.1\% & 51.2\% & 26.5\% & 45.9\% \\
StarVLA-GR00T & 63.5\% & 51.3\% & 28.2\% & 47.6\% \\
\bottomrule
\end{tabular*}%
}
\newcommand{\ManiUnitLiteTable}{%
\begin{tabular*}{\linewidth}{@{}l@{\extracolsep{\fill}}rrrr@{}}
\toprule
Model & Original & Base & Joint & Overall \\
\midrule
StarVLA-PI & 70.1\% & 67.3\% & 26.8\% & 54.7\% \\
StarVLA-GR00T & 77.8\% & 64.0\% & 29.6\% & 57.1\% \\
StarVLA-Cosmos2 & 63.2\% & 59.0\% & 13.8\% & 45.3\% \\
$\pi_{0.5}$ & 80.0\% & 65.8\% & 53.0\% & 66.3\% \\
\bottomrule
\end{tabular*}%
}

\newcommand{\ManiUnitTableFormat}{%
  \normalsize
  \setlength{\abovecaptionskip}{\baselineskip}%
  \setlength{\belowcaptionskip}{\baselineskip}%
}

\newcommand{\TableBoundaryParameters}{%
\begin{table}[htbp]
\ManiUnitTableFormat
\centering

\caption{Default parameters for temporal boundary refinement.}
\label{tab:boundary_parameters}
\setlength{\tabcolsep}{9pt}
\renewcommand{\arraystretch}{1.12}
\begin{tabular}{@{}lr@{}}
\toprule
Parameter & Value \\
\midrule
Monitored arm joints & First 3 per arm \\
Reference window & 500 frames \\
Smoothing window & 5 frames \\
Minimum joint departure & $15^{\circ}$ \\
Minimum qualifying duration & 5 frames \\
Gripper position-change threshold & $0.002\,\mathrm{m}$ \\
Gripper protection margin & $5\,\mathrm{s}$ per boundary \\
Minimum retained fraction & 0.2 \\
\bottomrule
\end{tabular}
\end{table}
}

\newcommand{\TableDatasetComparison}{%
\begin{table}[!htbp]
\ManiUnitTableFormat
\centering

\setlength{\tabcolsep}{4pt}
\renewcommand{\arraystretch}{1.15}
\caption{Comparison of manipulation datasets and benchmarks in terms of mobility, training annotations, and skill evaluation.}
\label{tab:dataset_comparison}
\begin{tabularx}{\linewidth}{@{}l*{5}{>{\centering\arraybackslash}X}@{}}
\toprule
\multirow{2}{*}{Dataset} & \multirow{2}{*}{Mobile} & \multicolumn{2}{c}{Training} & \multicolumn{2}{c}{Evaluation} \\
\cmidrule(lr){3-4}\cmidrule(lr){5-6}
& & \shortstack{Skill\\labels} & \shortstack{Language\\instr.} & \shortstack{Independent\\skill eval.} & \shortstack{Robot-state\\perturb.} \\
\midrule
\shortstack[l]{LIBERO\\\citep{liu2023libero}} & $\times$ & $\times$ & $\checkmark$ & $\times$ & $\times$ \\[3pt]
\shortstack[l]{CALVIN\\\citep{mees2022calvin}} & $\times$ & $\checkmark$ & $\checkmark$ & $\checkmark$ & $\times$ \\[3pt]
\shortstack[l]{RoboTwin 2.0\\\citep{chen2026robotwin}} & $\times$ & $\times$ & $\checkmark$ & $\times$ & $\times$ \\[3pt]
\shortstack[l]{RoboCasa365\\\citep{nasiriany2026robocasa365}} & $\checkmark$ & $\checkmark$ & $\checkmark$ & $\times$ & $\checkmark$ \\[3pt]
\shortstack[l]{BEHAVIOR-1K\\\citep{li2023behavior}} & $\checkmark$ & $\checkmark$ & $\checkmark$ & $\times$ & $\times$ \\
\midrule
ManiUnit (ours) & $\checkmark$ & $\checkmark$ & $\checkmark$ & $\checkmark$ & $\checkmark$ \\
\bottomrule
\end{tabularx}
\vspace{3pt}
\begin{minipage}{\linewidth}
\normalsize
$\checkmark$: provided; $\times$: not provided in the compared dataset or evaluation protocol.
Skill labels are temporally localized annotations within demonstrations.
Independent skill evaluation combines intermediate-state restoration with local success conditions for the corresponding operation.
Robot-state perturbation varies the initial robot base pose or joint configuration.
\end{minipage}
\end{table}
}

\newcommand{\TablePickUpResults}{%
\begin{wraptable}{r}{0.54\linewidth}
\vspace{-\intextsep}
\ManiUnitTableFormat
\vspace{\dimexpr-\abovecaptionskip-0.5pt\relax}
\centering
\caption{Pick Up success rates for Original and three scene variations:
target position (Pos.), target color, and distractor presence (Dist.).}
\label{tab:pickup_scene_variations}
\normalsize
\centering
\setlength{\tabcolsep}{2pt}
\renewcommand{\arraystretch}{1.45}
\begin{tabular*}{\linewidth}{@{}l@{\extracolsep{\fill}}rrrr@{}}
\toprule
Model & Original & Pos. & Color & Dist. \\
\midrule
StarVLA-PI & 68.6\% & 40.0\% & 60.0\% & 62.9\% \\
StarVLA-GR00T & 71.4\% & 20.0\% & 65.7\% & 62.9\% \\
StarVLA-Cosmos2 & 60.0\% & 22.9\% & 54.3\% & 60.0\% \\
$\pi_{0.5}$ & 91.4\% & 74.3\% & 80.0\% & 80.0\% \\
\bottomrule
\end{tabular*}
\end{wraptable}
}

\newcommand{\TableLongTasks}{%
\begin{wraptable}{r}{0.50\linewidth}
\vspace{-\intextsep}
\ManiUnitTableFormat
\vspace{\dimexpr-\abovecaptionskip+0.4pt\relax}
\centering
\caption{\textbf{Long-horizon task execution.}
Full-task success rates and mean Q-scores for the task policy alone and
the combined system, which uses a planner to coordinate the task and
skill policies. Overall averages the results across Turning On Radio
and Picking Up Trash.}
\label{tab:long}
\normalsize
\setlength{\tabcolsep}{1pt}
\renewcommand{\arraystretch}{1.45}
\begin{tabular*}{\linewidth}{@{}l@{\extracolsep{\fill}}rrrr@{}}
\toprule
 & \multicolumn{2}{c}{Task policy}
 & \multicolumn{2}{c}{Combined system} \\
\cmidrule(lr){2-3}\cmidrule(l){4-5}
Activity & \shortstack{Full-task\\success} & \shortstack{Mean\\Q-score}
 & \shortstack{Full-task\\success} & \shortstack{Mean\\Q-score} \\
\midrule
Radio & 4.0\% & 4.0\% & 24.0\% & 24.0\% \\
Trash & 4.0\% & 18.0\% & 12.0\% & 26.0\% \\
Overall & 4.0\% & 11.0\% & 18.0\% & 25.0\% \\
\bottomrule
\end{tabular*}
\end{wraptable}
}

\newcommand{\TableFullSkills}{%
\begin{widetable}[!t]
\ManiUnitTableFormat
\centering
\setlength{\tabcolsep}{3.1pt}
\caption{Per-skill success rates on ManiUnit-Full (\%).
Base and Joint denote Base Perturbation and Joint Perturbation.}
\label{tab:full_skills}
\begin{tabular*}{\linewidth}{@{\extracolsep{\fill}}lrrrrrrrr@{}}
\toprule
& \multicolumn{4}{c}{StarVLA-PI} & \multicolumn{4}{c}{StarVLA-GR00T} \\
\cmidrule(lr){2-5}\cmidrule(lr){6-9}
Skill & Original & Base & Joint & Mean & Original & Base & Joint & Mean \\
\midrule
Attach & 21.7\% & 11.7\% & 3.3\% & 12.2\% & 26.7\% & 1.7\% & 0.0\% & 9.4\% \\
Chop & 41.7\% & 31.7\% & 46.7\% & 40.0\% & 43.3\% & 31.7\% & 41.7\% & 38.9\% \\
Close Door & 83.3\% & 83.3\% & 50.0\% & 72.2\% & 90.0\% & 83.3\% & 46.7\% & 73.3\% \\
Close Drawer & 86.7\% & 95.0\% & 35.0\% & 72.2\% & 98.3\% & 98.3\% & 46.7\% & 81.1\% \\
Close Lid & 73.3\% & 66.7\% & 31.7\% & 57.2\% & 75.0\% & 61.7\% & 35.0\% & 57.2\% \\
Hang & 0.0\% & 0.0\% & 1.7\% & 0.6\% & 0.0\% & 1.7\% & 5.0\% & 2.2\% \\
Ignite & 100.0\% & 78.3\% & 8.3\% & 62.2\% & 95.0\% & 75.0\% & 5.0\% & 58.3\% \\
Insert & 23.3\% & 8.3\% & 10.0\% & 13.9\% & 26.7\% & 3.3\% & 1.7\% & 10.6\% \\
Open Door & 78.3\% & 75.0\% & 35.0\% & 62.8\% & 81.7\% & 83.3\% & 40.0\% & 68.3\% \\
Open Drawer & 36.7\% & 30.0\% & 5.0\% & 23.9\% & 51.7\% & 38.3\% & 10.0\% & 33.3\% \\
Open Lid & 51.7\% & 13.3\% & 20.0\% & 28.3\% & 45.0\% & 23.3\% & 15.0\% & 27.8\% \\
Pick Up & 48.3\% & 56.7\% & 28.3\% & 44.4\% & 65.0\% & 61.7\% & 36.7\% & 54.4\% \\
Place & 91.7\% & 70.0\% & 46.7\% & 69.4\% & 90.0\% & 68.3\% & 38.3\% & 65.6\% \\
Pour & 55.0\% & 43.3\% & 10.0\% & 36.1\% & 70.0\% & 53.3\% & 5.0\% & 42.8\% \\
Press & 45.0\% & 21.7\% & 6.7\% & 24.4\% & 56.7\% & 21.7\% & 1.7\% & 26.7\% \\
Release & 100.0\% & 100.0\% & 86.7\% & 95.6\% & 100.0\% & 91.7\% & 95.0\% & 95.6\% \\
Spray & 90.0\% & 85.0\% & 60.0\% & 78.3\% & 90.0\% & 85.0\% & 71.7\% & 82.2\% \\
Sweep & 85.0\% & 66.7\% & 25.0\% & 58.9\% & 85.0\% & 56.7\% & 33.3\% & 58.3\% \\
Turn Off Switch & 55.0\% & 56.7\% & 21.7\% & 44.4\% & 46.7\% & 56.7\% & 16.7\% & 40.0\% \\
Turn On Switch & 88.3\% & 80.0\% & 25.0\% & 64.4\% & 86.7\% & 76.7\% & 43.3\% & 68.9\% \\
Wipe & 6.7\% & 1.7\% & 0.0\% & 2.8\% & 10.0\% & 3.3\% & 3.3\% & 5.6\% \\
\bottomrule
\end{tabular*}
\end{widetable}
}

\newcommand{\TableLiteSkills}{%
\begin{widetable}[!t]
\ManiUnitTableFormat
\centering
\setlength{\tabcolsep}{3.1pt}
\caption{Per-skill success rates on ManiUnit-Lite (\%).
Base and Joint denote Base Perturbation and Joint Perturbation.}
\label{tab:mini_skills}
\begin{tabular*}{\linewidth}{@{\extracolsep{\fill}}lrrrrrr@{}}
\toprule
& \multicolumn{3}{c}{StarVLA-PI} & \multicolumn{3}{c}{StarVLA-GR00T} \\
\cmidrule(lr){2-4}\cmidrule(lr){5-7}
Skill & Original & Base & Joint & Original & Base & Joint \\
\midrule
Close Door & 95.0\% & 97.5\% & 45.0\% & 95.0\% & 90.0\% & 42.5\% \\
Open Door & 80.0\% & 96.7\% & 13.3\% & 93.3\% & 83.3\% & 10.0\% \\
Pick Up & 68.6\% & 77.1\% & 8.6\% & 71.4\% & 80.0\% & 14.3\% \\
Place & 91.4\% & 74.3\% & 80.0\% & 85.7\% & 57.1\% & 68.6\% \\
Pour & 22.9\% & 5.7\% & 8.6\% & 57.1\% & 25.7\% & 37.1\% \\
Press & 47.0\% & 31.0\% & 1.0\% & 42.0\% & 29.0\% & 9.0\% \\
Turn On & 85.7\% & 88.6\% & 31.4\% & 100.0\% & 82.9\% & 25.7\% \\
\midrule
\addlinespace[3pt]
& \multicolumn{3}{c}{StarVLA-Cosmos2} & \multicolumn{3}{c}{$\pi_{0.5}$} \\
\cmidrule(lr){2-4}\cmidrule(lr){5-7}
Skill & Original & Base & Joint & Original & Base & Joint \\
\midrule
Close Door & 62.5\% & 62.5\% & 22.5\% & 92.5\% & 92.5\% & 75.0\% \\
Open Door & 83.3\% & 80.0\% & 0.0\% & 100.0\% & 100.0\% & 63.3\% \\
Pick Up & 60.0\% & 54.3\% & 2.9\% & 91.4\% & 60.0\% & 71.4\% \\
Place & 97.1\% & 97.1\% & 57.1\% & 68.6\% & 51.4\% & 31.4\% \\
Pour & 0.0\% & 0.0\% & 0.0\% & 31.4\% & 20.0\% & 17.1\% \\
Press & 65.0\% & 36.0\% & 0.0\% & 76.0\% & 51.0\% & 27.0\% \\
Turn On & 74.3\% & 82.9\% & 14.3\% & 100.0\% & 85.7\% & 85.7\% \\
\bottomrule
\end{tabular*}
\end{widetable}
}

\newcommand{\TableSkillInventory}{%
\begin{table}[htbp]
\ManiUnitTableFormat
\centering

\caption{Skill-type inventory of ManiUnit-Full.}
\label{tab:skill_inventory}
\renewcommand{\arraystretch}{1.12}
\setlength{\tabcolsep}{10pt}
\begin{tabular}{@{}lrrr@{}}
\toprule
Skill type & Source activities & Subtasks & Segments \\
\midrule
Pick Up & 50 & 153 & 52,251 \\
Place & 45 & 184 & 49,574 \\
Chop & 6 & 18 & 7,624 \\
Open Door & 24 & 9 & 7,476 \\
Close Door & 20 & 7 & 6,048 \\
Pour & 7 & 14 & 2,993 \\
Sweep & 2 & 2 & 2,681 \\
Turn On Switch & 11 & 7 & 2,596 \\
Close Lid & 5 & 4 & 1,197 \\
Turn Off Switch & 3 & 3 & 1,000 \\
Spray & 2 & 2 & 804 \\
Open Lid & 3 & 3 & 794 \\
Insert & 1 & 2 & 600 \\
Release & 2 & 2 & 462 \\
Close Drawer & 1 & 1 & 400 \\
Open Drawer & 1 & 1 & 400 \\
Attach & 1 & 1 & 200 \\
Ignite & 1 & 1 & 200 \\
Press & 1 & 1 & 200 \\
Wipe & 1 & 1 & 200 \\
Hang & 1 & 1 & 199 \\
\midrule
\textbf{Total (21 skill types)} & \textbf{50} & \textbf{417} & \textbf{137,899} \\
\bottomrule
\end{tabular}
\end{table}
}

\begin{document}
\maketitle
\begingroup
\makeatletter
\let\@footnotetext\H@@footnotetext
\renewcommand{\thefootnote}{}
\footnotetext{This work was done during Guoting Wei's internship at IntelliFusion.}
\makeatother
\endgroup

\FigureOverview

\begin{abstract}
Long-horizon mobile manipulation requires a robot to navigate multi-room environments and execute a sequence of manipulation skills under a single natural language instruction.
Learning and evaluating these skills present three challenges: similar observations under a fixed task instruction may make skill selection ambiguous; even when a preceding skill succeeds, the robot state inherited by the next skill may deviate from its demonstrated starting states and affect execution; and task-level metrics hinder skill-specific diagnosis, while early failures leave later skills untested.
We therefore introduce ManiUnit, a manipulation skill dataset and benchmark built from 50 BEHAVIOR-1K activities.
Its dataset contains 137,899 segments across 21 skill types and 417 subtasks, and its benchmark contains 1,260 test instances.
Correspondingly, ManiUnit pairs each segment with an explicit subtask instruction; measures sensitivity to perturbations of the robot's starting base position or joint configuration; and restores intermediate simulator states and defines local success conditions so that each skill can be evaluated without executing preceding stages.
Evaluations of representative vision-language-action (VLA) policies show that similar aggregate scores can hide substantial per-skill differences.
The tested starting-state perturbations also degrade execution: on the full benchmark, joint perturbations reduce success rates by approximately 56\% relative to those from demonstrated starting states.
On two long-horizon activities, a skill policy trained on ManiUnit segments achieves 78.7\% local manipulation success, compared with 49.3\% for a task policy trained on complete demonstrations.
The trained skills further support complete-task execution on these activities, as coordinating the task and skill policies through a planner raises full-task success from 4.0\% to 18.0\%.

\end{abstract}

\section{Introduction}
\label{sec:intro}

Enabling robots to complete long-horizon mobile manipulation tasks in everyday environments is a central challenge for embodied AI.
Unlike tabletop manipulation, which operates in a fixed workspace, this setting requires robots to navigate across multi-room environments and execute a sequence of interdependent manipulation skills guided by a single natural language instruction.
Recent vision-language-action (VLA) models \citep{zitkovich2023rt2,kim2025openvla,black2025pi0,black2025pi05,nvidia2025gr00t,gemini2025robotics} have achieved strong results on tabletop benchmarks: StarVLA-$\pi$ \citep{starvla2026}, for example, reaches 95.7\% success on LIBERO \citep{liu2023libero} and 88.8\% under the randomized setting of RoboTwin 2.0 \citep{chen2026robotwin}.
Despite this progress, reliable execution of long-horizon tasks remains
challenging: the winning entry in the 2025 BEHAVIOR Challenge achieved
a full-task success rate of only 12.4\% on the held-out test set
\citep{larchenko2025task}, leaving substantial room for improvement.

Consider one such task, ``make microwave popcorn,'' which requires navigation interleaved with opening the microwave door, picking up the popcorn bag, placing it inside, closing the door, and turning on the microwave. Such sequences pose three challenges for skill learning and evaluation.
(1)~\textbf{Skill ambiguity}: when a policy follows the same task instruction throughout execution, it needs to infer the current stage from its observations and act accordingly.
This inference can break down when different stages share similar visual context: before loading food, the robot faces a closed microwave door and must open it, but after placing the food inside and closing the door, it again faces the same closed door yet must now turn on the microwave instead of opening the door.
(2)~\textbf{Starting-state mismatch}: during sequential execution, even a successfully completed skill may leave the robot in a base position or joint configuration that differs from the starting states in the next skill's training demonstrations, and this mismatch can cause a policy that succeeds from demonstrated starting states to fail when following a preceding skill.
(3)~\textbf{Evaluation opacity}: overall task success reflects both navigation and manipulation, making it difficult to identify which skills limit performance; moreover, an early failure can prevent later skills from being attempted.

These challenges motivate treating individual manipulation operations as units of training and evaluation.
We therefore introduce ManiUnit, a manipulation skill dataset and benchmark built by segmenting long-horizon BEHAVIOR-1K demonstrations \citep{li2023behavior} into such operations.
Figure~\ref{fig:overview} maps each challenge to the corresponding design choice.
\textbf{For skill ambiguity}, ManiUnit attaches an explicit subtask instruction to each segment, naming the operation, the objects it acts on, and the spatial context, keeping the intended operation explicit during skill learning and evaluation.
\textbf{For starting-state mismatch}, ManiUnit evaluates each skill
under three conditions: the demonstrated state, a base-position
perturbation, and a joint-configuration perturbation.
These controlled tests measure how variations in the robot's
starting state affect skill execution.
\textbf{For evaluation opacity}, ManiUnit constructs test instances by restoring intermediate simulator states and defining local success conditions for each subtask, so every skill can be evaluated independently without first executing the stages that precede it.
The dataset covers 50 activities with 21 skill types, 417 subtasks, and 137,899 segments with a mean duration of 17.9 seconds, each paired with head and wrist RGB streams, robot states, and actions.
The benchmark contains 1,260 test instances, drawn from 20 operation starts per skill type under the three starting-state conditions.

Evaluating representative VLA policies on ManiUnit shows that similar aggregate scores can hide substantial differences in per-skill performance.
The evaluations also reveal pronounced sensitivity to the tested starting-state perturbations: on the full benchmark, joint perturbations reduce success rates by approximately 56\% relative to those from demonstrated starting states.
On Turning On Radio and Picking Up Trash, we compare task and skill policies initialized from the same pretrained weights.
The skill policy trained on ManiUnit segments achieves 78.7\% local manipulation success, compared with 49.3\% for the task policy trained on complete demonstrations.
To deploy these skills in complete tasks, a planner coordinates the task and skill policies and supplies the appropriate subtask instructions (Figure~\ref{fig:overview}, bottom).
On the same activities, the combined system achieves 18.0\% full-task success, compared with 4.0\% for the task policy alone.

Our contributions are threefold.
(1) We introduce the ManiUnit dataset: 137,899 segments from 50 BEHAVIOR-1K activities with explicit subtask instructions, organized by skill type and subtask to support mixed-skill training and targeted data selection.
(2) We construct the ManiUnit benchmark: 1,260 test instances from restored intermediate states with local success conditions and separate base and joint perturbations, enabling independent per-skill evaluation and measurement of starting-state sensitivity.
(3) We evaluate representative VLA policies to characterize skill performance and starting-state sensitivity, and show on two long-horizon activities that ManiUnit-trained skills improve local manipulation success and support complete-task execution through policy composition.

\section{ManiUnit Dataset and Benchmark}
\label{sec:maniunit}

\subsection{Overview}
\label{sec:dataset_overview}
\label{sec:dataset_statistics}

To study manipulation skills independently within long-horizon tasks, we construct ManiUnit from the 2025 Challenge demonstrations of 50 BEHAVIOR-1K activities~\citep{li2023behavior}.
ManiUnit comprises a dataset of manipulation segments paired with explicit subtask instructions and a benchmark for evaluating selected operations from restored starting states.
Figure~\ref{fig:pipeline} summarizes how the source demonstrations are processed into these training and evaluation resources.

The dataset distinguishes three levels of organization.
A \textbf{\emph{skill type} }identifies an operation category, such as Pour.
A \textbf{\emph{subtask}} is defined by an instruction specifying the operation, objects, and spatial context, such as ``pour the grated cheese from the tupperware onto the pizza dough.''
A \textbf{\emph{segment}} records one demonstrated execution of that subtask.
Grouping segments by skill type and subtask allows researchers to pool demonstrations across source activities or select data for particular operations.
\textbf{ManiUnit-Full} contains 21 skill types, 417 subtasks, and 137,899 segments.
We also curate \textbf{ManiUnit-Lite}, a subset of 10,588 segments across 7 skill types, to reduce the computational cost of repeated experiments.
Table~\ref{tab:skill_inventory} in Appendix~\ref{app:overview} reports the per-skill inventory of ManiUnit-Full.

For evaluation, we select operation starts from the source demonstrations and restore the corresponding robot and scene states.
For each selected subtask, we manually define a local BDDL goal that specifies the object states and relations required to complete the manipulation operation.
The restored state, subtask instruction, and local goal together form a test instance, allowing the operation to be evaluated without executing preceding task stages.
To measure sensitivity to starting-state variations, we construct three conditions for each selected start: Original, Base Perturbation, and Joint Perturbation.
The Full and Lite evaluation suites contain 1,260 and 186 test instances, respectively.
Sections~\ref{sec:dataset_construction} and~\ref{sec:benchmark_construction} describe the dataset and benchmark construction procedures.

\FigurePipeline

\subsection{Dataset Construction}
\label{sec:data}
\label{sec:construction}
\label{sec:dataset_construction}

As shown in Figure~\ref{fig:pipeline}, the source demonstrations contain RGB observations, robot states, and actions, together with temporal annotations of navigation and manipulation.
Each annotation records an action label, the participating objects, and the action's start and end frames.
However, these intervals do not always correspond to the manipulation segments needed for training. 
For example, pushing an object into reach and picking it up may be annotated separately despite contributing to the same Pick Up operation, while a recorded pickup interval may also include navigation before grasping or inactivity afterward.
To preserve complete manipulation operations while reducing surrounding navigation and inactivity, we extract and merge the relevant action intervals and refine their boundaries.
We then curate an explicit subtask instruction for each resulting segment.

\paragraph{Segment extraction and merging.}
We first extract manipulation intervals based on the annotated actions.
For each source activity, we manually specify which annotations to retain and apply this selection across its episodes.
For example, in the Picking Up Trash activity, we retain all annotated pickup and placement actions, each yielding an initial segment.
The recorded start and end frames of each retained annotation provide the initial segment boundaries.

Since supporting motions may be annotated separately from the operation they serve, we define merging rules to combine them into complete manipulation segments.
In the board-game example in Figure~\ref{fig:pipeline}, pushing the game toward the bed edge prepares it for grasping.
We therefore combine the push and subsequent pickup into one Pick Up segment.
The merged segment spans the start of the first contributing action through the end of the last, retaining all intervening frames.
Appendix~\ref{app:extraction_rules} details the extraction and merging rules.

\paragraph{Temporal boundary refinement.}
We refine the boundaries of the extracted and merged segments to reduce navigation and inactivity around the target operation.
Reaching for and handling objects often require raising or extending the arms, so we use deviations from a fixed lowered-arm reference posture as a heuristic cue for manipulation.
We measure this deviation as
\begin{equation}
d_t=\max_{j\in\mathcal J}
\left|\bar q_{t,j}-q_j^{\mathrm{ref}}\right|,
\label{eq:arm_deviation}
\end{equation}
where $\mathcal J$ denotes the monitored joints of both arms, $\bar q_{t,j}$ is the smoothed position of joint $j$ at frame $t$, and $q_j^{\mathrm{ref}}$ is its reference position.
To filter out brief fluctuations, we identify intervals in which $d_t$ remains above a threshold for a minimum number of consecutive frames.
We retain the continuous span from the start of the first such interval to the end of the last.
To preserve grasp and release motions, we also detect gripper events from changes in finger-joint positions. We adjust the segment boundaries to retain the detected gripper events and a short interval before the first event and after the last.
We also apply checks against excessive trimming and manually inspect selected segment videos to assess the resulting boundaries.
Appendix~\ref{app:boundary_refinement} provides refinement details.

\paragraph{Instruction curation.}
The extracted segments retain their action labels and participating objects from the source annotations.
We group segments by their action and object combination.
For each group, we convert the simulator object identifiers to readable names and define an instruction template that specifies the action and the objects' roles.
For example, Pour segments with grated cheese, tupperware, and pizza dough yield ``pour the grated cheese from the tupperware onto the pizza dough,'' explicitly naming what is transferred, its source, and its destination.
We then manually review every episode across all source activities, checking each segment's instruction against its video.
This review refines the description of the operation, objects, and spatial relations, and corrects errors inherited from the source annotations.

\paragraph{Data organization.}
The curated segments are grouped by skill type and subtask instruction, supporting pooled training across activities and targeted selection of particular operations.
Each segment contains its skill-type label, subtask instruction, three RGB streams from the head and both wrist cameras, and temporally aligned robot states and actions.
We retain each segment's source activity and episode, frame boundaries, and object identifiers.
These records enable benchmark construction to locate the operation's starting state and identify the corresponding simulator objects (Section~\ref{sec:benchmark_construction}).
Appendix~\ref{app:catalog} presents recorded examples with their subtask instructions.

\subsection{Benchmark Construction}
\label{sec:benchmark}
\label{sec:benchmark_construction}

The curated segments provide recorded demonstrations for training, but evaluating a policy on these operations requires running the policy in simulation.
For each selected operation, we therefore restore a reference starting state from its source demonstration so that the policy can begin without executing preceding task stages.
We also define a local goal that specifies when the operation is complete.
Together with the segment's curated subtask instruction, the restored state and local goal form a test instance:
\begin{equation}
e=(s_0,\ell,g),
\qquad
g:\mathcal S\rightarrow\{0,1\},
\label{eq:evaluation_instance}
\end{equation}
where $\mathcal S$ is the simulator state space, $s_0$ is the initial robot and scene state, and $\ell$ is the subtask instruction.
The goal function $g$ returns one when the required object states and relations hold, and zero otherwise.
To assess sensitivity to starting-state differences that may arise during sequential execution, we further construct variants of $s_0$ that perturb the robot's base position or joint configuration while keeping $\ell$ and $g$ fixed.

\paragraph{Initial state restoration ($s_0$).}
The reference starting state must preserve the effects of earlier task stages on the robot and scene.
For example, a Place operation may begin with the target object already in the robot's gripper.
We locate this starting point using the segment's source episode and start frame.
A single recorded frame may not describe the complete scene: the states of sleeping objects are not always recorded, and object additions and removals are stored separately.
We therefore start from the episode's saved scene configuration and replay the recorded states and transitions up to the selected frame.
Because the saved state may lack the simulator's assisted-grasp constraint, we recover it where needed so that held objects remain attached when physics resumes.
We also stabilize the restored state to reduce drift caused by residual velocities from the preceding motion.
Appendix~\ref{app:state_restoration} details the restoration procedure.

\paragraph{Local success conditions ($g$).}
Restoring the starting state allows an operation to be executed independently, but evaluating its outcome also requires a success condition for that operation.
The original task goal describes completion of the entire activity and does not directly specify when each intermediate operation has succeeded.
We therefore manually define a local goal $g$ for each evaluated subtask using the original task definition, the curated instruction $\ell$, and the source action annotations.
We express the required object states and relations as BDDL predicates \citep{srivastava2022behavior,li2023behavior}.
The segment's recorded object identifiers bind these predicates to the same simulator objects named in the instruction.

For example, placing a soda can $o$ in a trash can $c$ requires the can to be inside $c$ and no longer touching the robot $r$:
\begin{equation}
g_{\mathrm{place}}(s)
=
\operatorname{Inside}(o,c;s)
\land
\neg\operatorname{Touching}(o,r;s).
\label{eq:place_goal}
\end{equation}
The goal must also reflect the scope of the demonstrated operation.
For Sweep, the full-task goal requires clearing mud from the entire floor, while an individual segment addresses only one patch.
We therefore evaluate removal of the mud particles targeted by that segment, allowing other patches to remain.
Appendix~\ref{app:local_goals} details the local goal definitions and object bindings.
Section~\ref{sec:setup} specifies the temporal success rule and execution budget.

\paragraph{Robot state perturbations.}
During sequential execution, each skill begins from the robot configuration left by preceding operations.
Even when those operations succeed, the resulting configuration may differ from those represented in training and affect the next skill's execution \citep{gu2023multiskill}.
To measure sensitivity to this starting-state mismatch, we treat each selected operation start as an \emph{evaluation anchor} and construct three starting-state conditions for it.
Original uses the reference state restored from the source demonstration.
Base Perturbation translates the robot's base while preserving its orientation and the arm, torso, and gripper configurations.
Joint Perturbation changes selected arm or torso joint positions while preserving the base pose and gripper configuration.
The subtask instruction $\ell$, target identities, and local goal $g$ remain fixed across conditions.

Each perturbed state must preserve the preconditions for beginning the operation; a Place operation, for example, must still begin with the intended object in hand.
We therefore verify the intended configuration changes and reject variants that introduce unintended collisions or disrupt the required object relations.
Appendix~\ref{app:state_augmentation} describes these checks and the realized perturbation magnitudes.

\paragraph{Instance validation.}
Before accepting a test instance, we load it through the evaluation pipeline to verify the restored state, object bindings, and manipulation preconditions.
We also check that the local goal is initially unsatisfied, $g(s_0)=0$, and manually inspect the rendered starting scene.
Invalid instances are corrected or replaced and validated again.
Appendix~\ref{app:construction_records} details these checks.

\paragraph{Skill-specific tests.}
\label{sec:testconstruction}
Beyond robot-state variations, the same construction procedure supports targeted tests of how scene factors affect a chosen skill.
We demonstrate this with Pick Up by varying target position, target color, and the presence of a nearby distractor, one factor at a time.
The robot's starting configuration, target identity, subtask instruction, and local goal remain fixed across these variants.
For position changes, associated objects or contents are moved with the target as needed to preserve their spatial relations.
Appendix~\ref{app:scene_variations} describes the construction of these variants.

To extend these tests to another skill, researchers can select source demonstrations and a factor of interest, specify the object bindings and local goal, and reuse the restoration and validation procedures above.
The shared subtask definitions link each test instance to corresponding training segments, supporting targeted data selection and fine-tuning where evaluation reveals weaknesses.

\section{Policy Composition for Long-Horizon Tasks}
\label{sec:policy_composition}

Long-horizon mobile manipulation requires navigation, object manipulation, and coordination across successive task stages.
ManiUnit isolates the manipulation stages, providing segments with explicit subtask instructions for training and independent tests for evaluation.
A skill policy trained on ManiUnit can therefore focus on executing specified operations, but complete tasks also require navigation between operations and decisions about which one comes next.
A task policy trained on complete demonstrations provides control across these stages under a task-level instruction.
This suggests a simple composition: retain the task policy as the default controller and delegate selected manipulation stages to the skill policy.

This composition has two requirements.
First, the skill policy should perform the delegated operations more reliably than the task policy, a premise supported by the comparison in Section~\ref{sec:local_composition}.
Second, the skill policy must be invoked at the appropriate stage and receive the corresponding subtask instruction.
A planner provides this coordination, using robot-state and visual histories to decide when to delegate control and which instruction to supply.

This division separates manipulation learning from task-level control and coordination, giving each component a distinct target for improvement.
ManiUnit supports focused training and diagnosis of manipulation skills.
Future work can improve navigation through task-policy training and coordination through planner design, including task-progress estimation, subtask selection, and policy switching.
Complete-task evaluation can then assess whether improvements in these components lead to better overall performance.
Appendix~\ref{app:long} provides implementation details, and Section~\ref{sec:long} evaluates the combined system on complete tasks.

\section{Experiments}
\label{sec:experiments}
We evaluate representative VLA policies on ManiUnit to examine per-skill
performance and sensitivity to robot starting states and scene conditions
(Section~\ref{sec:results}).
We then compare skill-level and task-level training on Turning On Radio
and Picking Up Trash (Section~\ref{sec:local_composition}) and assess whether
the resulting skills improve complete-task execution through policy
composition (Section~\ref{sec:long}).

\begin{widefigure}[!t]
\centering
\begingroup
\setlength{\tabcolsep}{0.5pt}
\renewcommand{\arraystretch}{1.15}
\noindent
\begin{minipage}[t]{0.50\linewidth}
\centering
\begin{minipage}[c][1.40in][c]{\linewidth}
\centering\ManiUnitFullTable
\end{minipage}\par\vspace{1pt}
{\normalsize (a) Full: baseline results}
\end{minipage}\hfill
\begin{minipage}[t]{0.47\linewidth}
\centering
\begin{minipage}[c][1.40in][c]{\linewidth}
\centering\includegraphics[width=\linewidth]{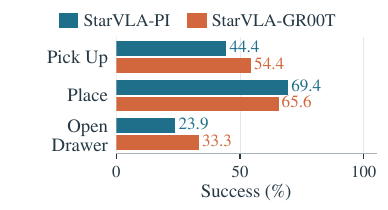}
\end{minipage}\par\vspace{1pt}
{\normalsize (b) Full: skill comparison}
\end{minipage}
\par\vspace{0.10in}\noindent
\begin{minipage}[t]{0.50\linewidth}
\centering
\begin{minipage}[c][1.40in][c]{\linewidth}
\centering\ManiUnitLiteTable
\end{minipage}\par\vspace{1pt}
{\normalsize (c) Lite: baseline results}
\end{minipage}\hfill
\begin{minipage}[t]{0.47\linewidth}
\centering
\begin{minipage}[c][1.40in][c]{\linewidth}
\centering\includegraphics[width=\linewidth]{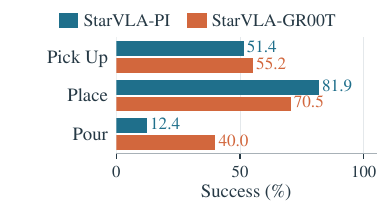}
\end{minipage}\par\vspace{1pt}
{\normalsize (d) Lite: skill comparison}
\end{minipage}
\par\endgroup

\caption{\textbf{Baseline performance on ManiUnit.}
(a) and (c) show success rates averaged equally across skill types
under each robot starting-state condition; Overall averages the three
conditions. (b) and (d) show the performance of PI and GR00T
on selected skills, averaged across conditions.}
\label{fig:benchmark_summary}
\end{widefigure}

\subsection{Experimental Setup}
\label{sec:setup}

\paragraph{Datasets and policies.}
We evaluate StarVLA-PI and StarVLA-GR00T~\citep{starvla2026}
on ManiUnit-Full and compare them with StarVLA-Cosmos2 and
$\pi_{0.5}$ on ManiUnit-Lite.
For each dataset, we pool segments across skill types and train
each model for five epochs.
StarVLA-PI and StarVLA-GR00T use pretrained
Qwen3.5-2B backbones~\citep{qwen2026qwen35}.
StarVLA-Cosmos2 uses a pretrained Cosmos-Predict2 backbone.
All three use randomly initialized action heads.
We fine-tune $\pi_{0.5}$ using RLinf~\citep{yu2025rlinf},
starting from policy weights pretrained on diverse multi-robot
data~\citep{black2025pi05}.
Appendix~\ref{app:training} provides the training configurations.

\paragraph{Skill evaluation.}
We evaluate 1,260 test instances on Full and 186 on Lite,
covering 21 and 7 skill types, respectively.
Both suites include three starting-state conditions.
Original tests whether the policy can execute each skill from
a starting state restored from a successful training demonstration.
During policy rollouts, even successful preceding skills may leave
the robot in a configuration different from the demonstrated one.
Base and Joint Perturbation therefore test starting-state sensitivity
by perturbing these same reference states.
A trial succeeds if the local goal holds for ten consecutive frames
within three times the subtask's mean demonstration duration.
We run each instance three times on Full and five times on Lite.
For each condition, we average success over repeated trials and
test instances within each skill type, then give all skill types
equal weight. Overall averages the three condition scores.
Appendix~\ref{app:statistics} reports per-skill results and
aggregation details.

\subsection{Skill-Level Evaluation}
\label{sec:results}

Figure~\ref{fig:benchmark_summary}(a) and (c) summarize policy
performance on Full and Lite.
On Full, StarVLA-PI and StarVLA-GR00T achieve similar overall
success rates of 45.9\% and 47.6\%, respectively.
On Lite, $\pi_{0.5}$ achieves the highest overall success rate
of 66.3\%, followed by StarVLA-GR00T (57.1\%),
StarVLA-PI (54.7\%), and StarVLA-Cosmos2 (45.3\%).
This advantage may partly reflect differences in pretraining:
$\pi_{0.5}$ starts from policy weights pretrained on diverse
multi-robot data~\citep{black2025pi05}, whereas the StarVLA
policies use randomly initialized action heads.

\paragraph{Per-skill performance.}
As shown in Figure~\ref{fig:benchmark_summary}(b) and (d),
results averaged across the three starting-state conditions
reveal different skill strengths among policies with similar
overall scores.
On Full, StarVLA-GR00T achieves higher success on Pick Up
(54.4\% vs.\ 44.4\%) and Open Drawer, while StarVLA-PI
performs better on Place.
On Lite, StarVLA-PI again achieves higher success on Place
(81.9\% vs.\ 70.5\%), but StarVLA-GR00T performs better
on Pour (40.0\% vs.\ 12.4\%).
These profiles identify different weaknesses in each policy
and can guide targeted data selection and training.

\paragraph{Robot starting states.}
As shown in Figure~\ref{fig:benchmark_summary}(a) and (c),
both robot-state perturbations reduce overall success rates
for all evaluated policies.
On Full, Base Perturbation reduces success relative to Original
by 14.8\% for StarVLA-PI and 19.3\% for StarVLA-GR00T.
Under Joint Perturbation, these relative reductions increase
to 55.9\% and 55.6\%, respectively.
All four policies on Lite also perform worse under both perturbations,
with larger drops under Joint Perturbation.
Under this condition, $\pi_{0.5}$ retains approximately two-thirds
of its Original success rate, compared with roughly 22--38\%
for the other three policies.
These results identify the robot's starting configuration as a major
factor in skill execution.
The smaller drop of $\pi_{0.5}$, which is pretrained on diverse
multi-robot data, suggests that broader coverage of starting states
during data collection or augmentation may improve robustness.

\TablePickUpResults

\paragraph{Object and scene conditions.}
\label{sec:targeted}
Holding the robot's starting configuration fixed, we examine
sensitivity to object and scene changes using Pick Up as a case study.
We vary target position, target color, and distractor presence
one at a time, while keeping the target identity, instruction,
and local goal unchanged.
Each of the four Lite policies is evaluated on seven test instances
per condition, with five repeats per instance.

As shown in Table~\ref{tab:pickup_scene_variations},
target-position changes produce the largest performance drop
for all four policies among the tested variations,
with relative reductions of approximately 19\% to 72\%
from their respective Original scores.
Color changes and distractors have smaller effects,
reducing success by at most \mbox{11.4 percentage points}.

\subsection{Skill-Level vs. Task-Level Learning}
\label{sec:local_composition}

We compare skill-level and task-level training on Turning On Radio
and Picking Up Trash to assess their effects on local manipulation
performance.
The task policy is trained on complete demonstrations with
task-level instructions.
The skill policy is trained jointly on the corresponding ManiUnit
segments across five object-specific subtasks, using their curated
subtask instructions.
Both policies are initialized from the same pretrained $\pi_{0.5}$
weights and trained for five epochs.
We evaluate both policies on the same 75 test instances, with five
instances per subtask under each of the three starting-state conditions.
Overall success averages the five subtasks and three conditions equally.

\begin{widefigure}[t]
\centering
\begingroup
\normalsize
\definecolor{ftbaseline}{HTML}{537A9E}
\definecolor{ftmixed}{HTML}{D18B55}
\setlength{\tabcolsep}{1.5pt}
\renewcommand{\arraystretch}{1.15}
\noindent
\begin{minipage}[t]{0.50\linewidth}
\centering
\begin{minipage}[c][1.60in][c]{\linewidth}
\centering%
\begin{tabular*}{\linewidth}{@{}l@{\extracolsep{\fill}}rrrr@{}}
\toprule
Policy & Original & Base & Joint & Overall \\
\midrule
\textcolor{ftbaseline}{Task policy} & 60.0\% & 56.0\% & 32.0\% & 49.3\% \\
\textcolor{ftmixed}{Skill policy} & 84.0\% & 76.0\% & 76.0\% & 78.7\% \\
\bottomrule
\end{tabular*}

\end{minipage}\par\vspace{1pt}
(a) Starting conditions
\end{minipage}\hfill
\begin{minipage}[t]{0.48\linewidth}
\centering
\begin{minipage}[c][1.60in][c]{\linewidth}
\centering\includegraphics[width=\linewidth]{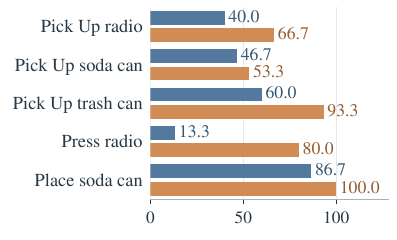}
\end{minipage}\par\vspace{1pt}
(b) Individual subtasks
\end{minipage}
\par\endgroup

\caption{\textbf{Skill-level vs. task-level learning.}
(a) shows overall success rates and success rates under each robot
starting-state condition for the task and skill policies.
(b) shows their success rates on the five subtasks, averaged across
starting-state conditions.}
\label{fig:skill_finetuning}
\end{widefigure}

As shown in Figure~\ref{fig:skill_finetuning}(a),
the skill policy achieves 78.7\% overall success, compared with
49.3\% for the task policy.
The advantage holds under all three starting-state conditions
and is largest under Joint Perturbation, where the skill policy
achieves 76.0\% success versus 32.0\% for the task policy.
Relative to Original, success drops by 8.0 percentage points
for the skill policy and 28.0 points for the task policy,
indicating greater robustness to the tested joint variations.

Figure~\ref{fig:skill_finetuning}(b) shows higher success for the
skill policy on all five subtasks when averaged across the three
starting-state conditions.
The largest gain is on Press radio, from 13.3\% to 80.0\%.
Success on Pick Up trash can also increases from 60.0\% to 93.3\%.

\subsection{Long-Horizon Task Execution}
\label{sec:long}

\TableLongTasks

We next test whether these local manipulation gains translate
into better complete-task performance.
Using the policies evaluated in Section~\ref{sec:local_composition},
we compare the task policy alone with the combined system
described in Section~\ref{sec:policy_composition}.
Both systems are evaluated on Turning On Radio and Picking Up Trash,
using the same ten episodes per activity and five repeats per episode.
We report full-task success and mean Q-score.
Q-score measures the fraction of task-goal conditions satisfied
at the end of each trial~\citep{li2023behavior}.
Appendix~\ref{app:long} provides implementation and evaluation details.

As shown in Table~\ref{tab:long}, the combined system increases
overall full-task success from 4.0\% to 18.0\% and mean Q-score
from 11.0\% to 25.0\%.
Full-task success increases from 4.0\% to 24.0\% on Turning On Radio
and from 4.0\% to 12.0\% on Picking Up Trash.
Figure~\ref{fig:planner_trajectory} in Appendix~\ref{app:long}
illustrates successful rollouts of both activities.
ManiUnit-trained skills can thus support complete-task execution
through policy composition.

\section{Conclusion}
\label{sec:conclusion}

We introduced ManiUnit, a manipulation skill dataset and benchmark
that treats individual operations within long-horizon tasks as units
of training and evaluation.
Its dataset reorganizes BEHAVIOR-1K demonstrations into manipulation
segments with explicit subtask instructions, while its benchmark
pairs restored simulator states with local goals to enable independent
evaluation under controlled starting-state variations.
Evaluations on ManiUnit support skill-level diagnosis and reveal
policy sensitivity to robot starting states.
On two activities, skill-level training improves local manipulation
success, and policy composition improves full-task success.

\paragraph{Limitations and future work.}
\label{sec:limitations}
Despite these gains, reliable skill execution remains challenging,
particularly under perturbed starting conditions.
Complete-task performance also depends on coordination decisions
that determine which subtask instructions to provide and when
to switch policies.
Future work can use ManiUnit's data and diagnostic evaluations
to improve individual skills.
In parallel, planners can be developed to better track task progress,
select subtasks, and determine appropriate switching times.
These directions allow manipulation learning and task coordination
to be studied separately, with their combined effectiveness
assessed through complete-task execution.

\bibliographystyle{maniunit_shortauthors}
\bibliography{references}

\clearpage
\appendix
\renewcommand{\topfraction}{0.92}
\renewcommand{\bottomfraction}{0.85}
\renewcommand{\textfraction}{0.08}
\renewcommand{\floatpagefraction}{0.78}
\setcounter{topnumber}{3}
\setcounter{bottomnumber}{2}
\setcounter{totalnumber}{5}
\setlength{\textfloatsep}{14pt plus 2pt minus 2pt}
\setlength{\floatsep}{10pt plus 2pt minus 2pt}
\setlength{\intextsep}{12pt plus 2pt minus 2pt}

\TableDatasetComparison

\section{Related Work}
\label{app:related}

\paragraph{Robot Demonstrations and Skill-Level Data Organization.}
Learning manipulation policies benefits from both broad demonstration coverage and explicit supervision for the operations within each demonstration.
Open X-Embodiment~\citep{collaboration2024openx} and DROID~\citep{khazatsky2024droid} expand coverage through aggregation across robot platforms and distributed data collection, respectively.
Beyond collection, MimicGen~\citep{mandlekar2023mimicgen} adapts demonstrations to new scene configurations, while SkillMimicGen~\citep{garrett2025skillmimicgen} composes adapted skill segments to generate new demonstrations.
Finer-grained annotations make the operations within these demonstrations explicit: NILS~\citep{blank2025nils} uses foundation models to segment demonstrations and assign language instructions, and RoboCOIN~\citep{wu2025robocoin} provides annotations at the trajectory, segment, and frame levels.
ManiUnit constructs manipulation segments from BEHAVIOR-1K demonstrations~\citep{li2023behavior} through action extraction and merging, boundary refinement, and manual instruction curation.
The resulting corpus groups segments across source activities by skill type and subtask, allowing researchers to pool data for training a shared skill policy or select data for a particular skill or subtask.

\paragraph{Skill Evaluation and Execution Robustness.}
Diagnosing long-horizon task performance requires understanding which operations fail and under what conditions.
LIBERO~\citep{liu2023libero} evaluates language-conditioned policies on tabletop manipulation tasks, while RoboTwin 2.0~\citep{chen2026robotwin} evaluates bimanual manipulation under domain randomization.
RoboCasa365~\citep{nasiriany2026robocasa365} includes atomic and composite kitchen tasks, and BEHAVIOR-1K~\citep{li2023behavior} provides long-horizon household activities in realistic simulation.
When a task combines multiple operations, overall success can obscure differences between skills, and early failures can leave later operations untested.
CALVIN~\citep{mees2022calvin} supports both independent task evaluation from restored demonstration states and sequential execution.
RoboEval~\citep{wang2026roboeval} and Eval-Actions~\citep{liu2026evalactions} provide complementary diagnostics of staged task progress and execution quality, respectively.
Robustness benchmarks examine changing execution conditions through environmental perturbations in Colosseum~\citep{pumacay2024colosseum} and language and visual variations in VLA-Arena~\citep{zhang2025vlaarena}.

ManiUnit combines independent skill evaluation with controlled variations in the robot's configuration at the start of an operation.
Restored intermediate states and local success conditions allow manipulation operations from long-horizon household demonstrations to be evaluated without executing preceding stages.
Perturbations of the robot's base position and joint configuration simulate variation in the starting configurations inherited from preceding operations and measure its effect on skill performance.
Table~\ref{tab:dataset_comparison} summarizes these capabilities.

\paragraph{VLA Policies and Hierarchical Control.}
Recent VLA models such as RT-2~\citep{zitkovich2023rt2}, OpenVLA~\citep{kim2025openvla}, and $\pi_0$~\citep{black2025pi0} map language instructions and visual observations to robot actions.
Under a single task instruction, however, a policy must infer the current operation, which can be ambiguous when different stages share similar visual context.
Hierarchical approaches provide explicit intermediate guidance.
SayCan~\citep{ichter2023saycan} selects skills by combining language-model estimates of usefulness with value-function estimates of execution feasibility, while Hi Robot~\citep{shi2025hirobot} supplies intermediate instructions through a high-level reasoner.
Other approaches integrate this guidance within the policy: $\pi_{0.5}$~\citep{black2025pi05} combines subtask prediction with conditioned action generation, and InstructVLA~\citep{yang2026instructvla} jointly trains embodied reasoning and action generation through instruction tuning.
ManiUnit supports instruction-conditioned manipulation by providing segments paired with manually curated subtask instructions for training a shared skill policy.
Section~\ref{sec:policy_composition} describes this coordination: a task policy handles navigation and transitions, while a planner decides when to invoke the skill policy and which subtask instruction to supply.

\section{Dataset Construction and Curation}
\label{app:curation}

This section provides further details on the dataset construction described in Section~\ref{sec:dataset_construction} of the main text, including skill segment extraction and merging rules, boundary-refinement parameters, and data consistency checks. 

\subsection{Skill Inventory and Coverage}
\label{app:overview}

Table~\ref{tab:skill_inventory} summarizes the source activities, subtasks, and segments associated with each skill type.
Subtask counts reflect the diversity of operations, while segment counts capture the number of recorded executions.
The mean segment duration is 17.9 seconds, computed from each segment's frame count and recorded frame rate.
Pick Up and Place appear in nearly all source activities and together account for 73.8\% of the segments, whereas task-specific operations such as Ignite and Wipe each come from a single activity.
Of the 21 skill types, 13 appear in multiple source activities, and 37 exact subtask instructions are shared across activities.
Organizing segments by skill type and subtask therefore supports targeted data selection and lets researchers pool demonstrations of the same operation across task contexts.
For evaluation, the Full benchmark assigns 20 anchors to each of the 21 skill types, giving every type equal representation regardless of its frequency in the training data (Section~\ref{sec:setup}).

We construct ManiUnit-Lite by selecting the eight source activities
with the highest task success rates in the 2025 BEHAVIOR Challenge.
These activities are Make Microwave Popcorn, Cook Hot Dogs,
Turning On Radio, Picking Up Trash, Moving Boxes to Storage,
Cook Bacon, Bringing Water, and Wash a Baseball Cap.
The curated manipulation segments from these activities form
a subset of 10,588 segments across seven skill types.

For the corresponding benchmark, we retain the test instances
from ManiUnit-Full whose evaluation anchors originate from
these eight activities.
This selects 186 of the 1,260 Full test instances, covering
62 anchors under Original, Base Perturbation, and Joint Perturbation.
Together, these dataset and benchmark subsets reduce the
computational cost of repeated experiments.

\TableSkillInventory

\subsection{Skill Segment Extraction and Merging}
\label{app:extraction_rules}

ManiUnit builds on the skill annotations provided with the BEHAVIOR-1K demonstrations.
Each source annotation records an action label, the objects involved, and the action's start and end frames.
We use these annotations to extract manipulation segments from full-task trajectories.
For the 50 source activities, we manually define 312 extraction rules and 48 merging rules.

Extraction rules specify which annotated actions to retain from each activity.
The same action label may appear several times within an episode for different objects.
To identify the intended operations, the rules use object information and execution order where needed.
Action labels are matched exactly, distinguishing annotations such as \emph{place on} and \emph{place on next to}.
For example, an episode may contain several pickup actions involving different objects.
The rules specify which pickups to retain, and each retained pickup becomes a separate segment.
The recorded start and end frames provide the initial boundaries of each extracted interval.
Merging rules combine supporting actions with the main operations they serve when the source annotations record them separately.
Each rule specifies the actions to combine and one of two adjacency conditions.

The first condition requires consecutive annotations in the source demonstration.
Pushing a board game to the edge of a bed and picking it up, for example, are annotated as two consecutive actions, although the push prepares the object for grasping.
The corresponding rule merges these two intervals into one Pick Up segment.
The second condition requires consecutive entries in the sequence of extracted manipulation operations, allowing navigation between a supporting action and the main operation.
Before hanging a picture, for example, the robot may transfer it between its hands, move to the hanging location, and then hang it.
The source annotations can therefore record navigation between the handover and hanging actions.
Because navigation is not extracted as a separate manipulation operation, the handover and hanging intervals become consecutive in the extracted sequence when no other selected operation lies between them.
The corresponding rule combines them into one Hang segment, beginning with the handover and ending with the hanging action, while retaining the navigation recorded in between.

\subsection{Temporal Boundary Refinement}
\label{app:boundary_refinement}

Boundary refinement locates the manipulation phase within each extracted interval using the arm-posture departure defined in Section~\ref{sec:dataset_construction}.
We monitor the first 3 joints of each arm and smooth their positions with a length-preserving moving average using zero padding.
The fixed reference posture is the mean of the smoothed joint positions over 500 lowered-arm frames from source activity 0, episode 1.
A consecutive run qualifies when its departure from this reference meets the threshold for at least the minimum duration.
We then retain the continuous interval from the start of the first qualifying run to the end of the last, including intervening pauses.
Table~\ref{tab:boundary_parameters} lists the defaults.

\TableBoundaryParameters

Two safeguards limit excessive trimming.
To preserve motions around grasp and release, we detect gripper events from the maximum absolute change in finger-joint positions between consecutive frames, using the unsmoothed positions of both grippers.
The candidate interval is extended as needed to include a margin before the first event and after the last, with both boundaries constrained to the original interval.
If the retained fraction falls below the specified threshold, we restore the full original interval.
Intervals with no qualifying arm run are retained for manual inspection.
This procedure shortens 13,664 segments in the released corpus.
We also inspect selected segment videos to check that the refined boundaries preserve the intended operation.

\subsection{Quality Control and Data Consistency}
\label{app:quality_control}

Each exported segment records its source activity and episode, together with its frame boundaries before and after refinement.
We clip the action annotations to the retained interval and reindex them relative to the segment, preserving their source indices for traceability.

We check the correspondence between segment identifiers and instruction records across the full corpus and find no mismatches.
For sampled segments, we also compare video frame counts, the lengths of robot-state and action sequences, and their source-frame indices.
These checks confirm that the inspected videos, states, and actions cover the same interval of the source demonstration.

\section{Benchmark Construction and Validation}
\label{app:testconstruction}

This section expands on the benchmark construction in Section~\ref{sec:benchmark_construction}.
We describe how source simulation records are used to restore each operation's starting state, how local goals are defined, and how test variants are constructed and checked.

\subsection{Initial State Restoration}
\label{app:state_restoration}

Evaluating a manipulation skill independently requires restoring the robot and scene to the conditions at the operation's start in the source demonstration.
We identify this point using the segment's source episode and frame index.
The restoration procedure addresses three requirements: recovering the scene history leading up to the operation, restoring any existing grasp, and reducing residual motion before policy execution.

\paragraph{Replaying the scene history.}
A single recorded frame may not contain all the information needed to reconstruct the scene.
The recording pipeline can omit the states of sleeping objects, relying on their previously recorded states during playback.
Object and particle-system additions and removals are stored separately in a transition history.
We therefore start from the episode's saved scene configuration and replay the recorded states and transitions up to the selected frame, collecting the information needed to restore the operation's starting conditions.

\paragraph{Recovering assisted-grasp constraints.}
Some operations begin with the robot already holding an object.
The simulator maintains this attachment through an \emph{assisted-grasp constraint}, which may be absent from the saved state even when the robot and object poses are correct.
Without this constraint, the object may slip when physics resumes.
We therefore check the grasp relation after restoration.

\paragraph{Stabilizing residual motion.}
A state captured during a demonstration may retain velocities from the preceding motion, causing the robot or scene objects to drift after restoration.
We reduce this residual motion before policy execution.
By default, the action-based generator uses 25 settling steps followed by 60 hold-action steps.
During the hold phase, the controllers maintain the robot configuration, command zero base velocity, and preserve the recorded gripper commands.
Velocity clearing further reduces drift.
Finally, we check that the restored state satisfies the operation's preconditions and that the local goal is not already satisfied before evaluation begins.

\subsection{Local Goal Definitions and Object Binding}
\label{app:local_goals}

The source BDDL goals describe full-activity outcomes and do not directly specify when individual intermediate operations are complete.
We therefore define a local goal for each evaluated subtask using the original task definition, the subtask instruction, and the source action annotations.
Each goal specifies the required object states and relations through BDDL predicates bound to the corresponding simulator instances.
For example, placing a soda can in a trash can requires the designated can to be inside the designated container and no longer touching the robot, as in Equation~\ref{eq:place_goal}.

Two skill types require additional handling because the operation either replaces objects or affects only a local region.

\paragraph{Chop.}
Cutting can replace the source object with pieces that have new simulator identities: a whole pepper, for example, becomes pepper slices once the cut is complete.
The local goal therefore tracks both the source object and the required pieces.
When the source object disappears, completion requires the corresponding pieces to be present and the remaining goal conditions to hold.
Its disappearance alone does not establish that cutting has succeeded.

\paragraph{Sweep.}
The full-task goal requires the entire floor to be clean, but an individual sweep segment targets one patch.
We identify the particles removed by the demonstrated operation and evaluate only that set.
To determine which particles the segment actually removes, we compare source-action replay with a reference rollout in which the robot holds still for the same duration: a particle belongs to the target set if it disappears during the sweep but remains present when the robot does nothing.
We store the selected particle identities with the test instance and use the same target set across all policy evaluations, so success requires removing those specific particles rather than clearing mud elsewhere.

\subsection{Robot State Perturbations}
\label{app:state_augmentation}

During long-horizon task execution, preceding skills may leave the robot in a configuration that differs from the demonstrated starting state of the next operation.
To measure sensitivity to these differences, we construct three starting-state conditions for each anchor while keeping the subtask instruction, target identities, and local goal fixed.
Original uses the reference state restored from the source demonstration.
Base Perturbation translates the robot's base while preserving its orientation and the arm, torso, and gripper configurations.
Joint Perturbation changes selected arm or torso joint positions while preserving the base pose and gripper configuration.

A perturbed state must preserve the conditions needed to begin the operation.
For a Place anchor, for example, the robot must still be holding the intended object after the perturbation.
We therefore check each candidate for unintended collisions, disrupted grasp relations, and displacement of contents relative to held containers.
We also verify that components intended to remain fixed have not changed beyond the allowed tolerances.
Candidates that fail these checks are rejected.

In Full, base translations range from 8.0 to 11.8\,cm, with a median of 9.9\,cm.
For Joint Perturbation, we measure the largest absolute joint change in each anchor; the median across anchors is 0.916\,rad.

\subsection{Skill-Specific Test Extensions}
\label{app:scene_variations}

The benchmark construction procedure also supports targeted tests of how scene factors affect individual skills.
We demonstrate this with Pick Up, varying target position, target color, and distractor presence separately.
Each variant keeps the robot's starting configuration, target identity, subtask instruction, and local goal fixed, allowing comparison with Original to assess sensitivity to the selected factor.

Color variants change the target's color through an object-specific texture override while preserving its geometry and pose.
Distractor variants add a nearby object from a different category.
The distractor receives a separate simulator identity, so the instruction and local goal continue to refer to the intended target.
Both variants use the Original state as their starting point.

Position variants require additional care when the target contains or supports other objects.
Moving a popcorn bag, for example, also requires moving its popcorn particles to preserve their arrangement within the bag.
The generator applies the same rigid transformation to the container and its associated particles, using their states from the same source frame.
This preserves their spatial relations at the modified starting state.
Each variant records its construction settings.

To extend these tests to another skill, researchers select source demonstrations and a scene factor of interest, specify the corresponding object bindings and local goal, and reuse the restoration and validation procedures described in this section.
The shared subtask definitions also link test instances to matching training segments for targeted data selection and fine-tuning.

\subsection{Test Records and Validation}
\label{app:construction_records}

Each exported test instance records its source episode and frame indices, subtask instruction, local goal, and construction settings, linking it back to the source operation and allowing it to be regenerated.
Before final release, we load every instance through the evaluation pipeline for a final validation pass.
This checks state restoration, object bindings, and manipulation preconditions, and verifies that the local goal is not already satisfied at the starting state.
We also manually inspect the rendered starting scene of every test instance to catch restoration errors or unintended scene changes that automated checks may miss.
Instances that fail either check are corrected or replaced.

\section{Experimental Details and Additional Results}
\label{app:implementation}

This section provides the training configurations, complete per-skill results, and details of the supplementary studies reported in the main text.

\subsection{Training Configurations}
\label{app:training}

Benchmark policies are trained for 5 epochs on their corresponding ManiUnit-Full or ManiUnit-Lite training sets, with one policy per model trained jointly across all skill types in the set.
Full evaluates StarVLA-PI and StarVLA-GR00T; Lite additionally evaluates StarVLA-Cosmos2 and $\pi_{0.5}$.
Each skill prompt specifies the skill type and full subtask instruction, for example: \emph{Task Skill Type: pick up. Task Prompt: pick up the radio from the coffee table.}

\paragraph{StarVLA.}
We largely follow the training configurations provided by StarVLA \citep{starvla2026}.
For StarVLA-PI and StarVLA-GR00T, we initialize the vision-language
backbone from Qwen3.5-2B~\citep{qwen2026qwen35} and randomly
initialize the action heads.
Policies predict chunks of 30 actions, each with 23 dimensions, and use a global batch size of 512.
Training uses AdamW with $\beta=(0.9,0.95)$, 5,000 warm-up steps, gradient clipping at 1.0, and a cosine learning-rate schedule.
Learning rates are $10^{-5}$ for the backbone and $10^{-4}$ for the action head.

StarVLA-Cosmos2 loads its text encoder, visual VAE, and
world-model transformer from the Cosmos-Predict2-2B-Video2World
checkpoint.
The text encoder and VAE remain frozen.
We jointly train the transformer and a randomly initialized
GR00T-style flow-matching action head.

We adapt the camera inputs for Cosmos2 and the joint-control interface for StarVLA.
The upstream Cosmos2 wrapper interprets an image list as a temporal sequence, whereas BEHAVIOR provides 3 simultaneous camera views.
We therefore combine a $224\times224$ head image with two $112\times112$ wrist images placed side by side below it, forming a $336\times224$ spatial mosaic (height $\times$ width).
This layout presents all 3 views as a single observation.
Resizing and padding preserve their aspect ratios, and training and inference use the same layout.

For joint control, we express the 14 arm-joint targets and the first 3 torso-joint targets as offsets from the joint positions observed at the start of each predicted chunk.
The reference positions remain fixed throughout the chunk.
Base velocities, the fourth torso joint, and gripper commands retain their native representation.
At inference, the adapter adds the reference positions back after denormalization to recover absolute joint targets.

\paragraph{$\pi_{0.5}$.}
We use the RLinf implementation \citep{yu2025rlinf}, with training settings informed by Openpi Comet \citep{bai2025openpi}.
The policy uses absolute joint targets and a global batch size of 256.
Training uses AdamW with a learning rate of $2.5\times10^{-5}$, $\beta=(0.9,0.95)$, 1,000 warm-up steps, and a cosine schedule.
Sampling follows the recorded frame distribution, and the random seed is 42.

\subsection{Per-Skill Success Rates}
\label{app:statistics}

In long-horizon task execution, a single task-level score cannot identify which specific skills are unreliable or how sensitive each is to starting-state variations.
An early failure may also prevent later skills from being attempted at all, leaving their performance unknown.
ManiUnit addresses both limitations by evaluating each skill independently from a restored starting state with its own local goal, and by testing each skill under three starting-state conditions to measure how robot configuration at entry affects execution.

Tables~\ref{tab:full_skills} and~\ref{tab:mini_skills} report per-skill success rates under Original, Base Perturbation, and Joint Perturbation on ManiUnit-Full and ManiUnit-Lite, respectively.
For each anchor and starting-state condition, we average success over repeated trials.
We then average across anchors within each skill type and give skill types equal weight when computing the score for each condition.
Overall success averages the three condition scores.

\TableFullSkills
\TableLiteSkills

\subsection{Planner and Complete-Task Evaluation}
\label{app:long}

This section provides implementation and evaluation details for the policy composition scheme in Section~\ref{sec:policy_composition}.
The combined system uses the task and skill policies compared in Section~\ref{sec:local_composition}.
Both policies are initialized from the same pretrained $\pi_{0.5}$ weights and trained for five epochs on their respective datasets.
The task policy uses complete demonstrations of Turning On Radio and Picking Up Trash, while the skill policy uses ManiUnit segments from these activities with skill-level instructions.
During execution, the task policy provides default control for navigation and transitions between stages.
The planner decides when to invoke the skill policy and which subtask instruction to supply.

\paragraph{Planner.}
The planner first distinguishes navigation from manipulation so that the skill policy is invoked during the stages it was trained to execute.
Robot-state histories capture arm, torso, and gripper configurations and their changes over time, providing cues about the current execution stage.
We therefore train a navigation classifier on these histories to estimate whether the robot is navigating or manipulating.

Identifying the specific operation also requires visual information about the scene and objects.
We use two complementary estimates to reduce premature switching on an uncertain prediction.
A subtask classifier combines robot-state history with head-camera features to predict the skill type.
Separately, a video-retrieval module matches recent observations to a database built from annotated training demonstrations of the two activities.
Each database entry associates video features with a subtask label and its object-specific instruction.
The classifier and retrieval module must agree on the skill type before the planner invokes the skill policy.
Both classifiers use separate task-conditioned causal Transformers with 8 layers, hidden width 128, and 4 attention heads.
Each robot-state observation is projected to a 128-dimensional token, and each classifier processes up to 32 observations sampled at 1\,Hz.
The subtask classifier additionally attends to frozen SigLIP features from the 8 most recent head-camera images.
The retrieval module uses a VideoMAEv2 encoder adapted on BEHAVIOR-1K videos to extract features from recent head- and wrist-camera clips.

The retrieval module accepts a subtask instruction when its confidence meets a predefined threshold.
The planner activates the skill policy when the navigation classifier identifies manipulation and the subtask classifier predicts the same skill type as the accepted instruction, with both classifiers meeting their respective confidence thresholds.
Otherwise, the task policy retains control.

Once a skill is active, the planner uses a lower subtask-confidence threshold to reduce interruptions.
If confidence falls below this threshold, a fixed grace period allows it to recover before control returns to the task policy.
This tolerance applies only while the navigation decision remains manipulation and the skill type and target instruction remain unchanged.
All policy switches occur at action-chunk boundaries.

\paragraph{Evaluation.}
We evaluate the composed system and the task-policy baseline on Turning On Radio and Picking Up Trash.
Each activity has 10 test episodes, each repeated 5 times, yielding 50 trials per activity for each system.

We report success rate and final Q-score.
Success rate is the proportion of trials that satisfy the complete-task BDDL goal.
Final Q-score measures partial task completion as the fraction of goal conditions satisfied at the end of a trial, using the best-matched goal clause \citep{li2023behavior}.
We compute success rate and mean final Q-score for each activity and give the two activities equal weight in the overall results.
Table~\ref{tab:long} reports both metrics as percentages.
Figure~\ref{fig:planner_trajectory} illustrates successful rollouts with their policy selections and subtask-instruction transitions.

\FigurePlannerTrajectory

\clearpage
\section{Visual Catalog of Manipulation Skills}
\label{app:catalog}

Figures~\ref{fig:catalog_objects_one}--\ref{fig:catalog_tools_two} illustrate ManiUnit's 21 manipulation skill types through recorded examples paired with their subtask instructions.
We group the skills into three categories according to the type of interaction.
\emph{Object handling} includes moving, placing, attaching, and releasing objects, as well as transferring their contents.
\emph{Opening and closing} includes interactions with doors, drawers, and lids.
\emph{Activation and tool use} includes pressing buttons, operating switches, and igniting materials, as well as chopping, spraying, sweeping, and wiping.

Each row shows five examples of a skill type, together with their full subtask instructions.
When all five examples share the same instruction, we display it once below the images.

\begingroup
\makeatletter
\setlength{\@fptop}{0pt}
\makeatother
\FigureCatalogObjectsOne
\clearpage
\FigureCatalogObjectsTwo
\clearpage
\FigureCatalogArticulatedOne
\clearpage
\FigureCatalogArticulatedTwo
\clearpage
\FigureCatalogToolsOne
\clearpage
\FigureCatalogToolsTwo
\clearpage
\endgroup

\end{document}